\documentclass[lettersize,journal]{IEEEtran}
\usepackage{amsmath,amsfonts}
\usepackage{amssymb}
\usepackage[ruled]{algorithm2e} 
\usepackage{booktabs} 
\usepackage{threeparttable}
\usepackage[colorlinks]{hyperref}
\usepackage{pifont}
\usepackage{multirow}
\usepackage{multicol}
\usepackage{makecell}
\usepackage{color}
\usepackage[dvipsnames]{xcolor}
\usepackage{array}
\usepackage{textcomp}
\usepackage{stfloats}
\usepackage{url}
\usepackage{verbatim}
\usepackage{graphicx}
\usepackage{cite}

\usepackage[numbers,sort&compress]{natbib}

\usepackage{subfigure}

\begin{document}

\title{Rethinking Scale Imbalance in Semi-supervised Object Detection for Aerial Images}

\author{Ruixiang~Zhang,~
\and Chang Xu,~
\and Fang~Xu,~
\and Wen~Yang,~
\and Guangjun~He,~
\and Huai~Yu,~
\and Gui-Song~Xia

\thanks{This work has been submitted to the IEEE for possible publication. Copyright may be transferred without notice, after which this version may no longer be accessible.}
\thanks{R. Zhang, C. Xu, F. Xu, W. Yang, and H. Yu  are with the School of Electronic Information, Wuhan University, Wuhan, 430072 China.  \emph{E-mail: \{zhangruixiang, xuchangeis, xufang, yangwen, yuhuai\}@whu.edu.cn}}
\thanks{G. He is with the State Key Laboratory of Space-Ground Integrated Information Technology, Beijing, 100086, China.  \emph{E-mail: hgjun$\_$2006@163.com}}
\thanks{G. S. Xia is with the School of Computer Science, Wuhan University,
Wuhan 430072, China.  \emph{E-mail: guisong.xia@whu.edu.cn}}
\thanks{Manuscript received XXX, 2022; revised XXX.}}

\markboth{JOURNAL OF LATEX CLASS FILES, VOL. 14, NO. 8, AUGUST 2021}
{Shell \MakeLowercase{\textit{et al.}}: A Sample Article Using IEEEtran.cls for IEEE Journals}

\maketitle

\begin{abstract}

This paper focuses on the scale imbalance problem of semi-supervised object detection~(SSOD) in aerial images. Compared to natural images, objects in aerial images show smaller sizes and larger quantities per image, increasing the difficulty of manual annotation. Meanwhile, the advanced SSOD technique can train superior detectors by leveraging limited labeled data and massive unlabeled data, saving annotation costs. However, as an understudied task in aerial images, SSOD suffers from a drastic performance drop when facing a large proportion of small objects.  
By analyzing the predictions between small and large objects, we identify three imbalance issues caused by the scale bias, \textit{i.e.,} pseudo-label imbalance, label assignment imbalance, and negative learning imbalance. To tackle these issues, we propose a novel Scale-discriminative Semi-Supervised Object Detection (S$^3$OD) learning pipeline for aerial images. In our S$^3$OD, three key components, Size-aware Adaptive Thresholding (SAT), Size-rebalanced Label Assignment (SLA), and Teacher-guided Negative Learning (TNL), are proposed to warrant scale unbiased learning. 
Specifically, SAT adaptively selects appropriate thresholds to filter pseudo-labels for objects at different scales. SLA balances positive samples of objects at different scales through resampling and reweighting. TNL alleviates the imbalance in negative samples by leveraging information generated by a teacher model.
Extensive experiments conducted on the DOTA-v1.5 benchmark demonstrate the superiority of our proposed methods over state-of-the-art competitors. Codes will be released soon.

\end{abstract}


\section{Introduction}

\label{sec.intro}

\IEEEPARstart{O}{bject} detection in aerial imagery (ODAI) is a significant research field\cite{9643004,9776580,zhang2020grs,shi2020orientation,9924205,9654160}. Currently, deep learning-based methods dominate the field of aerial image object detection~\cite{ding2021object,xia2018dota,li2020object, yi2021oriented,ding2019learning,wang2020learning}. 
Supervised by manual annotations, these aerial object detectors can achieve superior performance. However, the annotation process is laborious, time-consuming, and expensive, limiting the quantity of labeled data. This dilemma is even more pronounced in aerial images, where the ubiquitous small and densely arranged objects introduce a great burden to the labeling process.  
To better leverage the precious labeled data and the relatively easily available unlabeled data, semi-supervised object detection~(SSOD) methods attract more and more attention~\cite{jeong2019consistency, sohn2020simple,liu2021unbiased,xu2021end,li2022pseco,zhou2022dense}. Using only one-tenth labels~\cite{wang2023consistent}, the SSOD method nowadays can achieve comparable performance to the fully-supervised method on the MS COCO benchmark~\cite{lin2014microsoft}.

\begin{figure}[!t]
\centering
\includegraphics[width=0.46\textwidth]{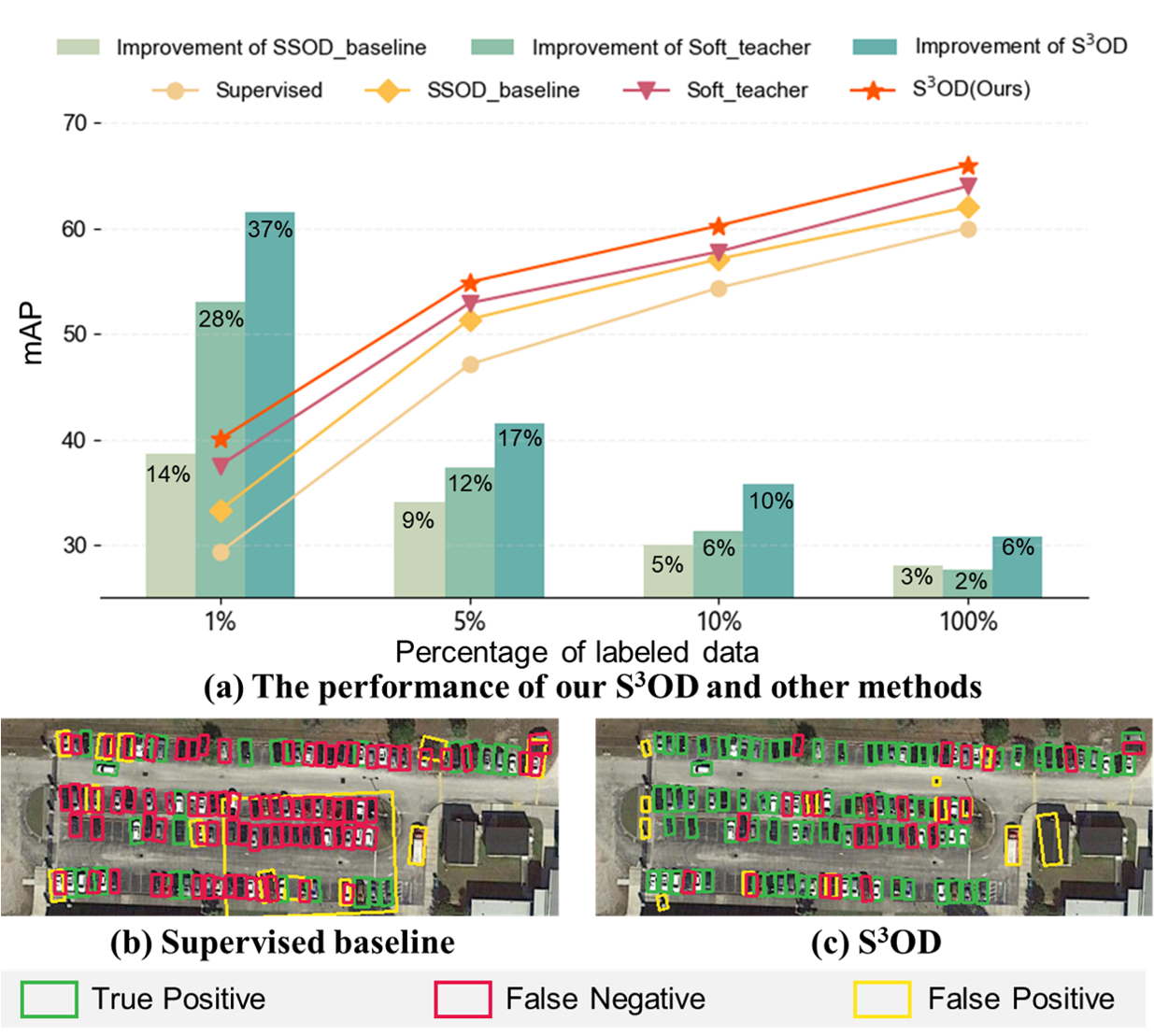}

\caption{(a) The line chart shows the comparison of SSOD-baseline,  Soft-teacher, and our S$^3$OD on DOTA-v1.5 validation set under different amounts of labelled training data. The bar chart shows the improvement over supervised baselines. (b) Detection results of supervised setting under 1$\%$ labeling rate. (c) Detection results of our S$^3$OD of semi-supervised setting under 1$\%$ labeling rate.}
\label{fig.1}
\end{figure}

In spite of the significant advancements in SSOD for natural images, we observe that they face an improvement bottleneck for aerial images. Some existing SSOD methods, such as STAC~\cite{sohn2020simple} and Soft-Teacher~\cite{xu2021end}, can yield performance improvements of more than 50$\%$ on MS COCO~\cite{lin2014microsoft} using only 1$\%$ labels. By contrast, when adapting these methods to aerial images, for example, the widely used DOTA-v1.5~\cite{xia2018dota}, as shown in Fig.~\ref{fig.1}, they provide fairly limited improvement. For Soft-teacher~\cite{xu2021end}, 105$\%$ on MS COCO~\cite{lin2014microsoft} \textit{v.s.} 28$\%$ on DOTA-v1.5~\cite{xia2018dota}.

\begin{figure}[!t]
\centering
\includegraphics[width=0.48\textwidth]{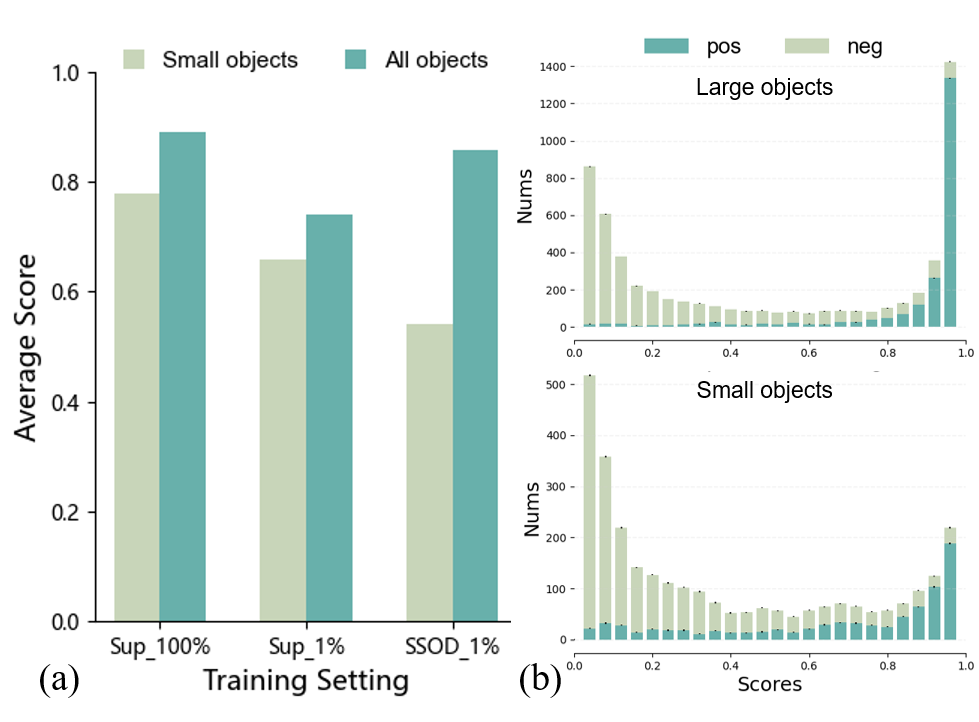}
\caption{(a) Comparing the average confidence scores of all objects and small objects in correct predictions under the three settings of supervised by 100$\%$ labeled data, supervised by 1$\%$ labeled data, semi-supervised by 1$\%$ labeled data. (The IoUs of the predicted boxes and GroundTruths greater than 0.5 are regarded as the correct predictions, noting that the average confidence score does not represent the test performance) (b) The distribution between the confidence scores and instances number of predictions during semi-supervised training. The top is for large objects, and the bottom is for small objects. (Note that on the vertical axis, the number of large objects is much higher than that of small objects.)}
\label{fig.2}
\end{figure}

This motivates us to rethink what factors lead to the dramatic degradation of SSOD methods on aerial imagery. For generality, we employ a vanilla teacher-student framework commonly used in the SOTA SSOD methods~\cite{wang2023consistent,chen2022dense,li2022pseco} for analysis. 
Specifically, observing that one of the typical characteristics of aerial imagery is its smaller object size, we investigate the indications of correctly predicted small objects\footnote{According to the definition in MS COCO~\cite{lin2014microsoft}, we empirically regard instances with a pixel area smaller than 1024 (32*32) pixels as small objects, the others with areas larger than 1024 pixels are called large objects.} versus all objects, as shown in Fig.~\ref{fig.2}~(a). 
Interestingly, we find out that the SSOD framework further enlarges the gap between the predicted confidence of small and large objects, which implies that the learning of small objects and large objects is severely unbalanced. Given the overwhelming ratio of small objects in aerial images (66\% in DOTA-v1.5~\cite{wang2021tiny}), it is thus natural to draw the conclusion that such scale bias in existing SSOD frameworks is one of the main obstacles impeding their performance for ODAI.

We identify three essential causes that lead to the scale bias by delving into the learning pipeline of existing SSOD frameworks, namely the pseudo-label imbalance, label assignment imbalance, and negative learning imbalance issues. 
First of all, pseudo-labels between small objects and large objects are unbalanced. In the general SSOD methods, a fixed pseudo-label threshold is often used to pick out high-quality predictions on unlabeled data as pseudo-labels, for example, the Soft-Teacher~\cite{xu2021end} chooses 0.9 to get the best performance. However, as shown in Fig.~\ref{fig.2}~(a), the average confidence of correct predictions of small objects tends to be much lower than larger objects, even for the fully supervised case. Therefore, a fixed threshold is bound to fail in simultaneously satisfying the optimal screening of different-sized objects. To further support this point of view, we elaborate on the confidence-sample number distribution of small/large objects during semi-supervised training, as shown in Fig.~\ref{fig.2}~(b). Compared to large objects, the correctly predicted small objects are more scattered across the whole confidence score ranges. The above observation indicates that the fixed threshold pseudo-labeling manner suffers from a quality-quantity dilemma for different-sized objects. On the one hand, the high threshold in previous works can ensure high-quality pseudo-labels while leading to a drastic loss of supervision information for small objects. On the other hand, although lowering the threshold can compensate more pseudo-labels for small objects, it will introduce much low-quality and inaccurate supervision for large objects.

\begin{figure}[!t]
\centering
\includegraphics[width=0.425\textwidth]{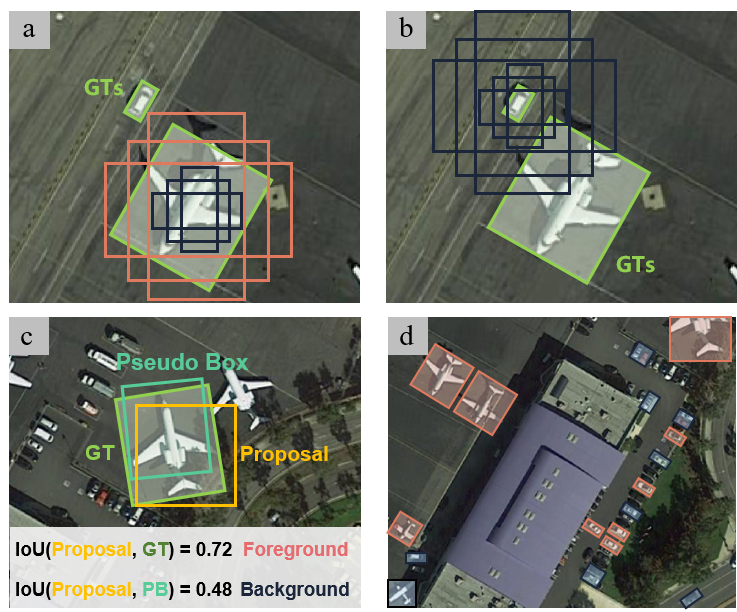}
\caption{The imbalance issues in the assignment of positive and negative samples in SSOD. The green represents GroundTruths or pseudo-labels used for supervision, the red represents the proposals or positive samples regarded as the foreground, and the dark blue represents the proposals or negative samples regarded as the background. (a) For large objects, more foreground proposals can be assigned. (b) For small objects, it is more challenging to match foreground proposals. (c) Biased pseudo-labels may misclassify the foreground proposals into the background. (d) Sparse pseudo-labels lose a lot of true objects, which are likely to be picked as negative samples, confusing the detector.}
\label{fig.3}
\end{figure}

Moreover, small objects are confronted with sample insufficiency and lack of supervision problems during label assignment in the SSOD framework, exacerbating the unbalanced learning. To illustrate these problems, we take the mainstream two-stage anchor-based detector Faster R-CNN~\cite{ren2016faster} for analysis. The Faster R-CNN heuristically presents a set of anchors at each feature point in the RPN stage. For small objects, a slight location deviation between the ground truth (GT) and the anchor will lead to severe mismatch issues~\cite{xu2022detecting,xu2022rfla}, yielding low IoU~\cite{xu2023dynamic} between these two boxes' region. 
This characteristic of small objects brings about two adverse effects to the SSOD framework, we provide schematic diagrams in Fig.~\ref{fig.3}~(a), (b), and (c). From one perspective, heuristic-tuned anchors are easy to drift from small GTs, and small GTs are prone to be matched with much fewer positive anchors than large objects. In other words, small objects are facing a positive sample/anchor insufficiency issue. 
From another perspective, small objects are lacking in supervision during the training process. Since foreground small objects are harder to be matched with pseudo-labels, the average number of positive supervision (positive classification/regression loss) posed to each de facto small object is much smaller than the large object. Therefore, the SSOD is biased towards focusing on the learning of larger objects.

In addition, negative sample learning is inclined to mistakenly suppress foreground small objects. Here we show two pieces of evidence. First of all, as discussed in Fig.~\ref{fig.2}~(b), the confidence of correctly predicted small objects is dispersed across the whole score range. Second, the recall of small objects is much lower than large objects, there remain many undiscovered small objects in the ``background'' region.
Hence, when sampling negative samples for classification, the de facto foreground small objects are more likely to be wrongly treated as the background, as shown in Fig.~\ref{fig.3}~(d), depressing the recall.

To address the aforementioned issues in SSOD, we propose a novel framework called Scale-discriminative Semi-Supervised Object Detection (S$^3$OD) for aerial imagery. Built upon the popular SSOD framework, S$^3$OD introduces three new modules: Size-aware Adaptive Thresholding (SAT), Size-rebalanced Label Assignment (SLA), and Teacher-guided Negative Learning (TNL). These modules aim to optimize the performance of SSOD in aerial images from the perspectives of pseudo-label selection, positive sample assignment, and negative learning. Specifically, SAT performs the pseudo-labeling of small and large objects in a \textit{dividing-and-conquering} manner, allowing for the retention of more pseudo-labels for small objects, thereby alleviating the quantity imbalance between large and small pseudo-labels. 
SLA utilizes a distribution-based re-sampling strategy, mitigating the impact of positive sample quantity imbalance problems between small and large objects. Additionally, the label assignment strategy is modified by reweighting the loss, thus reducing the adverse influence of small objects' lack of supervision. 
TNL leverages the information from the teacher model to correct the selection of negative samples. For one, the teacher model helps filter out proposals that are likely to be positive samples during preliminary screening. For another, the predictions from the teacher model can further uncover hard samples, enhancing the network's discriminative capability.

Overall, the main contributions of this paper are as follows:
\begin{itemize}	
    \item We point out that the scale imbalance problem is one of the key obstacles impeding SSOD's performance on aerial imagery. We further systemically identify the pseudo-label imbalance, label assignment imbalance, and negative learning imbalance issues in a standard SSOD framework.
 
	\item We propose a novel SSOD framework for aerial images, called S$^3$OD. Specifically, we introduce Size-aware Adaptive Threshold (SAT) to select more appropriate pseudo-labels, Size-rebalanced Label Assignment (SLA) for balanced anchor assignment, and Teacher-guided Negative Learning (TNL) for discriminative negative learning.

	\item Extensive experiments on representative aerial image datasets demonstrate the significant improvements and advancements achieved by our proposed methods.
\end{itemize}

The rest of this paper is arranged as follows. First, we introduce related work about aerial image object detection and SSOD in Section II. Then, we provide a detailed description of our method in Section III, including the overall framework and the proposed three modules. The experiments and analysis are discussed in Section IV. Furthermore, we also discuss the insights and limitations in Section V. Finally, we draw a conclusion in Section VI.

\section{Related Work}
In this section, we briefly review the recent progress of object detectors for aerial images. Then we provide a brief review of the relevant SSOD methods. Finally, we review the SSOD methods for aerial images.

\subsection{Object Detection in Aerial Images}

In the past decade, object detection has been greatly advanced with the development of deep learning. The mainstream detection frameworks include a series of anchor-based detectors~\cite{ren2016faster,cai2018cascade,lin2017focal,qiao2021detectors}, anchor-free detectors~\cite{tian2019fcos,law2018cornernet,duan2019centernet}, and trendy transformer-based detectors~\cite{carion2020end,zhu2020deformable}.
These algorithms have achieved excellent results on natural images. Different from general natural images, one huge difference in ODAI is that the objects are arbitrary-oriented, so the horizontal bounding box cannot represent the object boundary well. In recent years, detectors based on rotated boxes occupy a dominant position in ODAI. For example, the Rotated RPN~\cite{ma2018arbitrary} tackles rotated object detection by employing additional rotated anchor boxes. RoI Transformer~\cite{ding2019learning} learns to convert RPN-generated horizontal proposals into rotated ones, while S2ANet~\cite{han2021align} utilizes the DCN to explicitly align features with anchors. Oriented R-CNN~\cite{xie2021oriented} generates oriented proposals directly from anchors at the RPN stage. Additionally, there are also several works~\cite{xie2021oriented,li2022oriented,li2021fcosr} that customize anchor-free detectors for rotated detection. 

Another main characteristic of the aerial image is its high proportion of small objects, bringing severe challenges for existing object detectors.
Super-resolution, as a straightforward and effective idea, is incorporated into small objects detection in several works~\cite{shermeyer2019effects,courtrai2020small,bashir2021small}. 
Besides, the SCRDet~\cite{yang2019scrdet} and SCRDet++~\cite{yang2022scrdet++} optimize the feature extraction process and emphasize richer feature information, to enhance small objects. 
Moreover, some works have found that the label assignment strategy in the existing detectors is extremely unfriendly to small objects, and proposed some new label assignment strategies like NWD~\cite{xu2022detecting}, RFLA~\cite{xu2022rfla}. 
DCFL~\cite{xu2023dynamic} also explores dynamic priors and balanced learning for small object detection. 

The above methods effectively optimize the detector scheme based on the characteristics of objects in aerial images, thereby enhancing detection performance. However, these deep learning-based methods usually require plenty of training data with expensive annotation costs. In contrast, we explore the application of semi-supervised detection algorithms for aerial images to effectively alleviate the demand for annotations.

{\subsection{Semi-Supervised Object Detection}}

The early semi-supervised learning methods in the computer community are developed for image classification. One key idea is consistency regularization~\cite{sajjadi2016regularization,laine2016temporal,tarvainen2017mean,berthelot2019mixmatch,berthelot2019remixmatch,xie2020unsupervised,lee2022contrastive}, which aims to force the network to produce consistent predictions on data with different augmentations. Another idea is pseudo-label learning~\cite{lee2013pseudo,grandvalet2004semi,xie2020self,rizvedefense,pham2021meta}, which uses self-training to obtain pseudo-labels for unlabeled samples through the pre-detector training with labeled data. Most SOTA methods combine the above two ideas~\cite{sohn2020fixmatch, zhang2021flexmatch, kim2021selfmatch, wang2022np, zheng2022simmatch,  chen2023boosting}. Inspired by the above methods, semi-supervised learning has also made a lot of progress in detection in recent years.

Nowadays, the dominant line of works~\cite{jeong2019consistency, sohn2020simple, jeong2021interpolation, zhou2021instant, wang2021data, li2022rethinking, zhang2022mind} combines the ideas of pseudo-label learning and consistency regularization into the framework of object detection. STAC~\cite{sohn2020simple} is one representative method, which uses labeled data to train the network, and then predicts and filters suitable pseudo-labels on unlabeled images with different augmentations. After that, all the data are mixed to train together. Then, some methods such as Unbiased teacher~\cite{liu2021unbiased}, Humble teachers~\cite{tang2021humble}, Interactive self-training~\cite{yang2021interactive}, Soft-teacher~\cite{xu2021end}, introduce the Exponential Moving Average~(EMA) idea in MeanTeacher~\cite{tarvainen2017mean} into semi-supervised object detection. The model is updated after each iteration of training, which realizes end-to-end training and also provides a new paradigm for subsequent SSOD works. Based on this paradigm, the current semi-supervised object detection research lays emphasis on how to select reliable pseudo-labels and how to use pseudo-labels for more effective supervised learning.

Regarding the selection of pseudo-labels, on one hand, some studies explore more robust metrics for measuring the quality of pseudo-labels, incorporating the accuracy of localization with classification confidence. For example, in Soft-teacher~\cite{xu2021end}, box jittering is employed to obtain the accuracy of the box, then boxes with high accuracy are used for regression supervision. Similarly, Unbiased Teacherv2~\cite{liu2022unbiased} and RUPL~\cite{choi2022semi} directly model regression uncertainty to select the pseudo-labels with high location accuracy. On the other hand, there are several works exploring different pseudo-label thresholds. S4OD~\cite{zhang2022s4od} defines the threshold based on F1-score on the validation set. ASTOD~\cite{vandeghen2022adaptive} selects a threshold that can ensure the recall. Consistent-Teacher~\cite{wang2023consistent} selects classwise thresholds to get better matching thresholds.

In terms of effectively exploiting pseudo-labels for learning, Soft-teacher~\cite{xu2021end} and PseCo~\cite{li2022pseco} leverages the predictions of teacher networks to assist pseudo-label supervision. PseCo~\cite{li2022pseco}, Consistent teacher~\cite{wang2023consistent}, and ARSL~\cite{liu2023ambiguity} propose to better assign positive and negative samples with the supervision of inaccurate pseudo-labels. YOLOv5~\cite{xu2023efficient} and LabelMatch~\cite{chen2022label} group pseudo-labels into reliable and unreliable ones, and apply different forms of supervision. DLS~\cite{chen2022dense} and Dense Teacher~\cite{zhou2022dense} supervise the student network via dense predictions from the teacher network.

From the perspective of consistency regularization, strategies such as feature alignment at different scales are employed in MixTeacher~\cite{liu2023mixteacher}, SED~\cite{guo2022scale}, and PseCo~\cite{li2022pseco}. MUM~\cite{kim2022mum} applies mixup at both the image and feature levels, introducing stronger perturbations to the samples to enhance consistency regularization.
Additionally, there are also explorations from the perspective of sampling. CropBank~\cite{zhang2022semi} constructs a sample bank to supplement samples from minority classes, mitigating the impact of class imbalance. Active Teacher~\cite{mi2022active} optimizes the sampling of unlabeled data by evaluating it from multiple dimensions and selecting appropriate unlabeled data for training. Furthermore, methods like Omni-DETR~\cite{wang2022omni} and Semi-DETR~\cite{zhang2023semi} explore semi-supervised detection frameworks based on the Transformer. 

The aforementioned methods have achieved promising results. However, the specific characteristics of aerial images, like small and arbitrary orientations, make limitations in achieving the best performance in semi-supervised detection tasks in the context of aerial images. In contrast, our method takes these characteristics into account and presents a suitable SSOD framework for aerial images.

\vspace{0.5em}
{\subsection{SSOD for Aerial Images}}

Indeed, there have been several studies focusing on SSOD in aerial images. \cite{chen2018semi} utilizes a GAN network to generate adversarial negative samples for training the network's classification. However, it has only been tested on sparse small datasets and lacks universality. SOOD~\cite{hua2023sood} adopts the current mainstream semi-supervised detection paradigm and incorporates the orientations of targets in aerial images into the semi-supervised detection algorithm, achieving performance improvement. This method strengthens supervision based on the rotation consistency of densely arranged targets in aerial images. However, it still does not effectively address the challenges posed by small objects, which remains an urgent problem to be tackled. 
In contrast, our method carefully considers the influence of small objects on different stages of SSOD in aerial images. We tackle the issue of imbalance caused by scale bias by selecting appropriate pseudo-labels and effectively utilizing them for supervision. In particular, we present very competitive results on SSOD for aerial images.

\begin{figure*}[!t]
\centering
\includegraphics[width=6.8in]{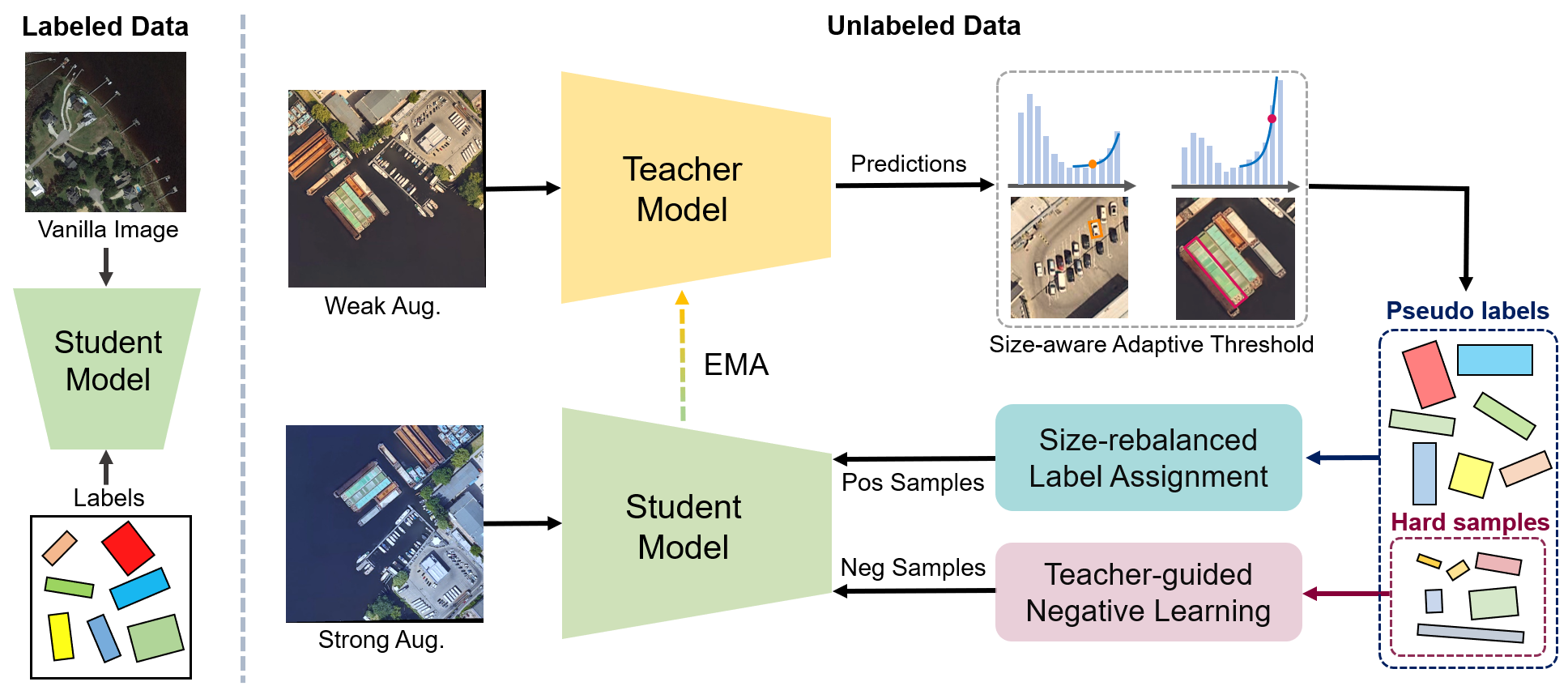}
\caption{The pipeline of the proposed S$^3$OD. Each training batch contains both labeled and unlabeled data. The labeled data uses Ground Truth for general supervised training. The training of unlabeled data is based on the teacher-student model, and the teacher network is updated by the EMA of the student network. The teacher network inferences the unlabeled data, and then selects appropriate pseudo-labels through SAT. SLA assigns positive samples based on the pseudo-labels. TNL strengthens the negative learning based on the ambiguous predictions of the teacher network.}
\label{fig.4}
\end{figure*}

\section{Methods}

This section details our proposed method S$^3$OD. First, we will introduce the overall architecture of the basic semi-supervised object detection we adopted, and then we will introduce our proposed methods for SSOD in aerial images, including SAT, SLA, and TNL.

\subsection{The Basic Framework} 
We follow the mainstream paradigm of pseudo-label learning for the overall framework~\cite{xu2021end, liu2021unbiased, zhou2022dense}, as shown in Fig.~\ref{fig.4}.
We take the common teacher-student model as the basic framework, where the student model is updated with the normal back-propagation, and the teacher model is the Exponential Moving Average (EMA) of the student model. 
Given the labeled dataset $D_l=\left \{ I_l,G_l \right \} $ and the unlabeled dataset $D_u=\left \{ I_u \right \}$, $I_l$ and $G_l$ represent the labeled images and the corresponding GTs, $I_u$ represent the unlabeled images. In each training iteration, labeled and unlabeled data are randomly sampled from $D_l$ and $D_u$, respectively, to train the student branch. 

For the labeled data, a standard training pipeline is employed. The labeled data images are input to the network for forward propagation, and the supervised loss $\mathcal{L}_{sup}$ is calculated by: 
\begin{equation}
\mathcal{L}_{sup}= \mathcal{L}_{det}(I_l,G_l).
\end{equation}

For the unlabeled data, weak and strong augmentations are respectively applied to the teacher and student models to enforce consistency regularization. The purpose of this step is to encourage the network to output consistent predictions for perturbed data and enhance the extraction of semantically invariant features. Following previous works, weak augmentation typically includes random flipping and resizing, while strong augmentation involves random rotation, translation, shearing, erasing, solarizing, adjusting color, contrast, sharpening, etc. In this paper, we did not explore new augmentation methods, and all strong and weak augmentation used are off-the-shelf approaches.
After applying strong and weak augmentations, each unlabeled image produces two views. The view with weak augmentation is input into the teacher model for inference, resulting in predicted detection results. Based on a predetermined threshold, suitable predictions are selected as pseudo-labels $P_u$ for the unlabeled image. The view with strong augmentation, along with the pseudo-labels $P_u$, is fed into the student model. By leveraging the supervision signal provided by the pseudo-labels, the unsupervised loss $\mathcal{L}_{unsup}$ is calculated by:
\begin{equation}
\mathcal{L}_{unsup}= \mathcal{L}_{det}(I_u,P_u).
\end{equation}

The overall loss function is formulated as: 
\begin{equation}
\mathcal{L}= \mathcal{L}_{sup}+\alpha * \mathcal{L}_{unsup},
\end{equation}
where $\alpha$ represents a weight parameter utilized to balance the contribution of unlabeled data.

To enable a fair comparison with previous methods, we take the well-established rotated version of Faster R-CNN as the basic detector. Without losing generality, our approach is applicable to any anchor-based detection method with the two-stage framework.

\subsection{Size-aware Adaptive Thresholding} 

Most of the existing SSOD methods utilize a fixed threshold to filter pseudo-labels. However, when processing data with unbalanced distribution, fixed thresholds are not always conducive to all situations. Previous works~\cite{xu2021end, liu2021unbiased} have pointed out the class confidence imbalance problem, where the class-wise reweighing or sampling strategy was correspondingly proposed to tackle this issue. Nevertheless, as analyzed in Sec.~\ref{sec.intro}, existing works commonly neglect the size-confidence imbalance issue in aerial imagery. 
Towards this end, we propose to define pseudo-labels in a size \textit{dividing-and-conquering} manner, and we name this strategy Size-aware Adaptive Thresholding (SAT).

In the SAT, we first compute the statistics of all predicted results from the teacher model with confidence scores higher than 0.5 within a batch. These results are then divided into two distributions based on the size of the object bounding boxes: one for large objects and the other for small objects. After this, the $P$-th percentile of each distribution is chosen as the threshold for large and small predictions, respectively, in order to retain a balanced quantity of reliable pseudo-labels for different-sized objects.

\subsection{Size-rebalanced Label Assignment}

After obtaining a series of pseudo-labels, we need to assign positive or negative (\textit{pos/neg}) labels to predefined anchors based on the obtained pseudo-labels to train the object detector. Meanwhile, the sample assignment between pseudo-labels and anchors plays quite a significant role in the performance of SSOD~\cite{li2022pseco, wang2023consistent}.
During this process, we observe two problems for aerial images, \textit{i.e.,} the sample insufficiency issue and the feature inconsistency issue, as discussed in Sec.~\ref{sec.intro}. 

To address these issues, we propose the Size-rebalanced Label Assignment~(SLA) strategy, which explores size-balanced learning through two key aspects: distribution-based re-sampling and size-aware re-weighting. In the distribution-based re-sampling, we model the rotated box $(cx, cy, w, h, \theta)$ into the two-dimensional Gaussian distribution $\mathcal{N}(\boldsymbol{\mu}, \boldsymbol{\Sigma})$~\cite{yang2021rethinking, yang2021learning}, in which the box's geometry center $\boldsymbol{\mu} = (cx, cy)$ serves as the Gaussian’s mean vector. And $\boldsymbol{\Sigma}$ is the covariance matrix of the Gaussian distribution, which can be computed by:

\begin{equation}\small
    \mathbf{\Sigma}=\begin{bmatrix}
        \cos{\theta} & -\sin{\theta} \\
        \sin{\theta} & \cos{\theta}
    \end{bmatrix}\begin{bmatrix}
    \frac{w^2}{4} & 0 \\ 
    0 & \frac{h^2}{4}
    \end{bmatrix}\begin{bmatrix}
        \cos{\theta} & \sin{\theta} \\
        -\sin{\theta} & \cos{\theta}
    \end{bmatrix}.
    \label{gaussian_model}
\end{equation}

Based on the distribution-based modeling, we use the Wasserstein distance to measure the similarity between pseudo-boxes and anchors since the Wasserstein distance is demonstrated conducive to tiny objects~\cite{xu2022rfla, xu2022detecting, yang2021rethinking} in label assignment, mainly owing to its ability to measure non-overlapping boxes. Specifically, the Wasserstein distance between the Gaussian pseudo-box $\mathcal{N}_p(\boldsymbol{\mu}_p, \boldsymbol{\Sigma}_p)$ and the Gaussian anchor $\mathcal{N}_a(\boldsymbol{\mu}_a, \boldsymbol{\Sigma}_a)$ has a closed form solution, which can be simplified as: 
\begin{equation}
\begin{aligned}
W_{2}^{2}(\mathcal{N}_p, \mathcal{N}_a) = & \left\|\mathbf{\mu}_p-\mathbf{\mu}_a\right\|_2^2 \\
& +\operatorname{Tr}\left(\boldsymbol{\Sigma}_p+\boldsymbol{\Sigma}_a-2\left(\boldsymbol{\Sigma}_p^{1 / 2} \boldsymbol{\Sigma}_a \boldsymbol{\Sigma}_p^{1 / 2}\right)^{1 / 2}\right).
\end{aligned}
\label{eq.gwd}
\end{equation}

For brevity, we call this metric WD. With WD, we can now calculate the similarity between all preset anchors and pseudo-boxes. For each pseudo-box, we assign the top $K$ anchors that yield the highest similarity with the pseudo-box as positive samples. In general, the strategy of WD with top $K$ sampling can alleviate the imbalance among the positive anchor sampling process to some extent. On the one hand, compared with IoU, distribution-based modeling has a wider definition domain and the WD can obtain the similarity score between a given pseudo-box and all anchor boxes in the image. On the other hand, the top $K$ sampling strategy can solve the problem that large objects are easier to be matched with more anchors~\cite{xu2022rfla}.

In addition to the distribution-based re-sampling, we design a novel size-aware re-weighting method. During the training process, we count the quantity of large and small positive samples in each batch. When calculating loss, we re-weight the loss of large and small positive samples to further balance the contribution of large and small objects in training. Given the loss of small positive samples $\mathcal{L}_s$, and loss of large positive samples $\mathcal{L}_l$, the re-weighted loss of all positive samples can be calculated by:
\begin{equation}
 \mathcal{L}_{pos}  = \frac{N_{pos}}{2N_s} \sum_{i=0}^{N_s} \mathcal{L}_s^i + \frac{N_{pos}}{2N_l} \sum_{j=0}^{N_l} \mathcal{L}_l^j ,
\end{equation}
where $N_s$ and $N_l$ are the number of small and large positive samples, $N_{pos} = N_s + N_l$ is the number of all positive samples.

\subsection{Teacher-guided Negative Learning}

To overcome the concern that negative sample learning will mistakenly suppress undiscovered small objects, we introduce the Teacher-guided Negative Learning~(TNL) approach for unbiased negative sample learning.

Object detectors' predictions are not always \textit{black and white}. More practically, there exist numerous ambiguous predictions that cannot certainly be labeled as foreground or background via a simple threshold. Due to limited pixels and information, this ambiguity is more severe for small object prediction.
Existing SSOD methods usually discard these ambiguous predictions, avoiding misleading gradients while sacrificing their potential contribution to training. Fortunately, some study on other semi-supervised learning tasks (e.g., U$^2$PL~\cite{wang2022semi}, ANL~\cite{chen2023boosting}) suggests that the ambiguous results predicted by the network can be excavated to aid learning. Thus, we cast a further look at ambiguous predictions in the SSOD task for ODAI.

In the proposed TNL, we design two steps to make the detector rethink the utility of these ambiguous samples. In the first step, the regions of the negative proposals are input into the teacher model to obtain classification confidence scores of background. Subsequently, we select negative proposals with higher background scores (larger than 0.7) for negative sample supervision, aiming to effectively eliminate False Negatives (FN). 
However, this approach gives rise to a new issue: the selected negative samples often turn out to be overly simplistic, discarding the learning of hard samples. This limitation impacts the network's discriminative capacity, leading to a significant increase in False Positives (FP) at the later stages of training. 
Hence, to enhance the network's discriminative ability of hard samples, we incorporate the network's predictions with confidence scores lower than 0.5 as soft hard samples into the negative sample proposals. Based on the original confidence predictions, we re-weight the loss of these hard samples, the lower the confidence score, the higher the weight, thus granting them greater significance during the training of negative samples. The loss $\mathcal{L}_{neg}$ of all negative samples can be calculated by:
\begin{equation}
\mathcal{L}_{neg}=\sum_{i=0}^{N_h} 2(1-s_i^2)\mathcal{L}_h^i+\sum_{i=0}^{N_n} \mathcal{L}_n^i,
\end{equation}
where $N_h$ and $N_n$ are the numbers of hard negative samples and normal negative samples, $N_{neg} = N_h + N_n$ is the number of all negative samples. $s$ are the confidence scores of the hard negative samples.

\section{Experiment}

\subsection{Dataset and Evaluation Protocol}

We use DOTA-v1.5 for experiments, which is a typical dataset in previous small object detection study~\cite{xu2023dynamic} and semi-supervised object detection study~\cite{hua2023sood}.
DOTA-v1.5 is updated based on DOTA-v1.0~\cite{DOTA_2018_CVPR}. Compared to v1.0, the images in v1.5 remains unchanged, but there are additional annotations for small objects. These enriched annotations of small objects allow the dataset to better reflect the characteristics of real-world aerial imagery objects.
The DOTA-v1.5 comprises 2,806 large-scale aerial images and 40,289 annotations. It is divided into three sets. The training set consists of 1,411 images, the validation set has 458 images, and the test set contains 937 images without released annotations. Following the common setting of SSOD~\cite{li2022pseco,xu2021end,liu2021unbiased}, we conducted two sets of experiments: \textbf{Partially Labeled Data} and \textbf{Fully Labeled Data}.

In the \textbf{Partially Labeled Data} experiments, 1$\%$, 5$\%$, and 10$\%$ of the training set are randomly sampled as labeled data, while the remaining data serves as unlabeled data. To mitigate the impact of random sampling, we perform 5-fold cross-validation for each sampling rate and reported the mean and variance of the results. 
In the \textbf{Fully Labeled Data} experiments, we utilize the test set without released annotations as unlabeled data, while using the entire training set as labeled data.

Note that aerial images are typically large-size and are often cropped before training. Moreover, sampling from the entire large images at a 1$\%$ sampling rate would make it difficult to cover all the categories adequately. Sampling from the cropped smaller images allows for a better representation of the overall data distribution. Therefore, in our experiments, we first perform cropping on all the images before sampling.

In all experiments, we report the evaluation results on the validation set using the commonly used mean Average Precision~(mAP) under the IoU threshold 0.5, following the MS COCO evaluation metric.

\begin{table*}[t!]
	\caption{Quantitative comparison between the proposed S$^3$OD and SOTA SSOD methods. Experiments are performed on the validation set of DOTA-v1.5 under the training with different labeling rates. The best results are in bold.}
	\label{tab:res}
	\resizebox{\textwidth}{!}{%
		\begin{tabular}{@{}cccccccc@{}}
			\toprule
			\multirow{2}{*}{Methods} & \multirow{2}{*}{Setting} & \multicolumn{3}{c}{Partially Labeled Data}                     & \multicolumn{1}{c}{Fully Labeled Data}                     \\ \cmidrule(l){3-6} 
			&                           & 1$\%$   & 5$\%$   & 10$\%$ & 100$\%$    \\ \midrule
			Rotated Faster RCNN \cite{lin2017focal}         & Supervised & 29.30$\pm$1.62 & 47.08$\pm$0.97 & 54.36$\pm$0.98 & 63.6   \\
			\midrule
			SSOD baseline        & Semi-supervised & 33.30$\pm$2.54 & 51.36$\pm$0.82 & 57.08$\pm$0.74 & 64.9 \textcolor{ForestGreen}{(+1.3)}  \\
			STAC~\cite{sohn2020simple}   & Semi-supervised & 33.98$\pm$2.13 & 50.86$\pm$1.63 & 56.72$\pm$0.75 & 63.8 \textcolor{ForestGreen}{(+0.2)} \\
			Unbiased Teacher~\cite{liu2021unbiased} & Semi-supervised & 33.46$\pm$2.35 & 48.20$\pm$1.14 & 54.74$\pm$0.84 & 64.8 \textcolor{ForestGreen}{(+1.2)} \\
			Soft Teacher~\cite{xu2021end}         & Semi-supervised & 37.52$\pm$2.44 & 52.90$\pm$1.24 & 57.78$\pm$0.32 & 65.1 \textcolor{ForestGreen}{(+1.5)}  \\
			PseCo~\cite{li2022pseco}         & Semi-supervised & 37.60$\pm$2.46 & 52.98$\pm$1.41 & 58.26$\pm$0.85 & 65.5 \textcolor{ForestGreen}{(+1.9)}  \\
			\textbf{S$^3$OD (Ours)}                   & Semi-supervised            & \textbf{40.02$\pm$1.68} & \textbf{54.88$\pm$1.22} & \textbf{60.24$\pm$0.42}    & \textbf{67.3 \textcolor{ForestGreen}{(+3.7)}} &  \\ \bottomrule
		\end{tabular}%
	}
\end{table*}

\begin{table*}[t!]
    \centering
	\caption{The performance of the proposed S$^3$OD and SOTA SSOD methods on several representative categories in the validation set of DOTA-v1.5. The training use 1$\%$ labeled data.  The best results are in bold.}
	\label{tab:cls}
	\resizebox{0.98\textwidth}{!}{%
		\begin{tabular}{@{}cccccccccc@{}}
			\toprule
			Methods         & Setting & SP & PL & SV & LV & BR & ST   & HB & mAP \\ \midrule
			Rotated Faster RCNN \cite{lin2017focal}         & Supervised & 60.9 & 68.7 & 22.6 & 44.9 & 12.1 & 43.6 & 28.2 & 29.4   \\
			\midrule
			SSOD baseline        & Semi-supervised & 60.1 & \textbf{89.3} & 9.1 & 50.9 & 21.4 & 50.8  & 18.5 & 36.0 \textcolor{ForestGreen}{(+6.6)}  \\
			STAC~\cite{sohn2020simple}   & Semi-supervised & \textbf{71.0} & 78.1 & 23.8 & 52.3  & 15.2 & 49.9  & 29.2 & 34.2 \textcolor{ForestGreen}{(+4.8)}\\
			Unbiased Teacher~\cite{liu2021unbiased} & Semi-supervised & 65.1 & 88.7 & 14.4 & 44.1 & 22.1 & 48.2 & 27.2 & 35.3 \textcolor{ForestGreen}{(+5.9)}\\
			Soft Teacher~\cite{xu2021end}         & Semi-supervised & 69.8 & 88.3 & 24.1 & 57.7  & 19.3 & 54.1 & 27.1 & 37.5 \textcolor{ForestGreen}{(+8.1)}\\
			PseCo~\cite{li2022pseco}         & Semi-supervised & 69.9 & 87.3 & 29.8  & 48.3 & 15.3 & 55.2  & \textbf{35.1} & 37.9 \textcolor{ForestGreen}{(+8.5)}\\
			\textbf{S$^3$OD (Ours)}    & Semi-supervised  & 70.9 & 88.5 & \textbf{39.7}    & \textbf{58.7}   & \textbf{23.6} & \textbf{68.6} & 29.0 & \textbf{41.9 \textcolor{ForestGreen}{(+12.5)}} \\ \bottomrule
		\end{tabular}%
	}
\end{table*}

\subsection{Implementation Details}
We use Rotated Faster RCNN~\cite{ren2016faster} as the base rotated object detector and choose ResNet50 w/ FPN as the backbone. The implementation of the base detector is based on the MMRotate framework~\cite{zhou2022mmrotate}. Following the settings in aerial image object detection, all large-sized images are cropped to $\rm{1024}\times\rm{1024}$ pixels with a 200-pixel overlap.

The supervised baseline is trained using stochastic gradient descent~(SGD) optimizer. We set the learning rate (LR) to 0.002 and the batch size (BS) to 4.
All SSOD methods are also trained using SGD optimizer with an LR of 0.002 and a BS of 5 (4 unlabeled images and 1 labeled image). The beta value is set to 4.0 to control the contributions of unlabeled data.

For partially labeled data, we train on 1 RTX4090 GPU for 100k iterations with 1$\%$ and 5$\%$ labeled data. For 10$\%$ labeled data, we train the model for 160k iterations on 1 RTX4090 GPU to ensure more comprehensive training.
For fully labeled data, we train the model for 300k iterations on 1 RTX4090 GPU.

\begin{figure*}[t!]
	\centering  
	\subfigure[Supervised~\cite{ren2016faster}]{
		\includegraphics[width=0.175\textwidth]{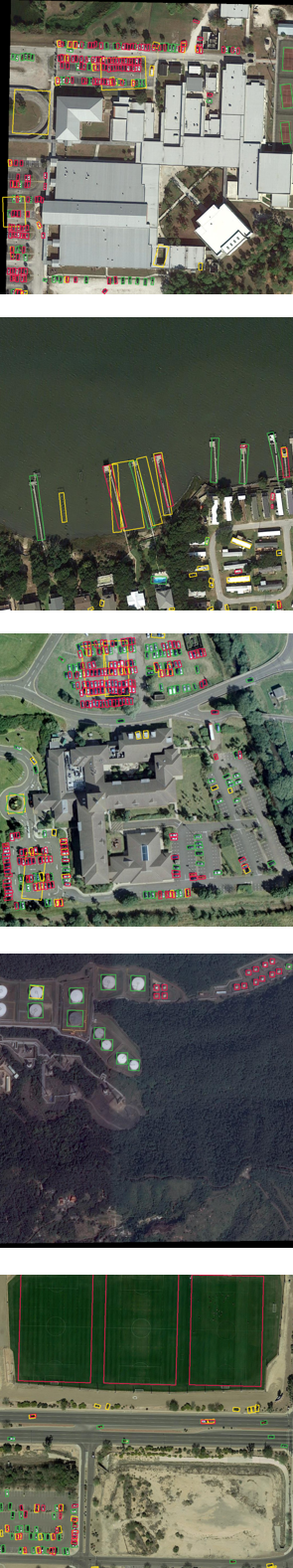}}
	\subfigure[Unbiased Teacher~\cite{liu2021unbiased}]{
		\includegraphics[width=0.175\textwidth]{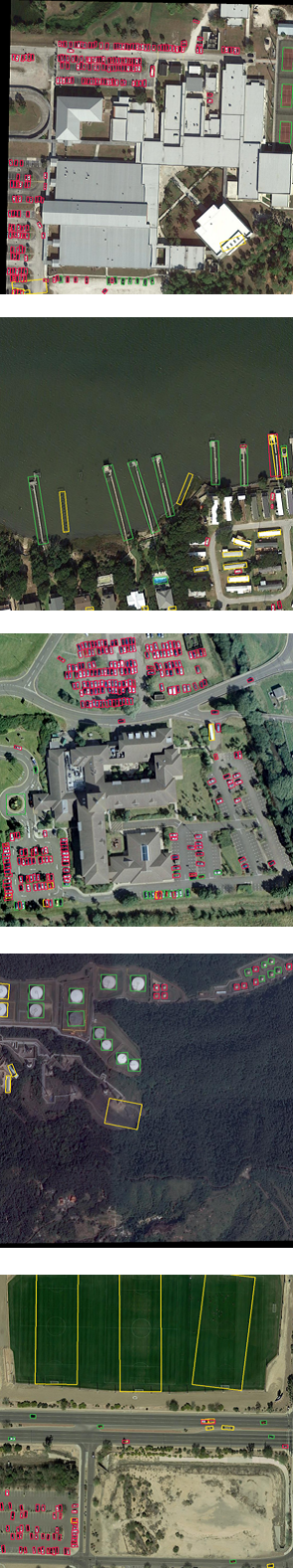}}
	\subfigure[Soft Teacher~\cite{xu2021end}]{
		\includegraphics[width=0.175\textwidth]{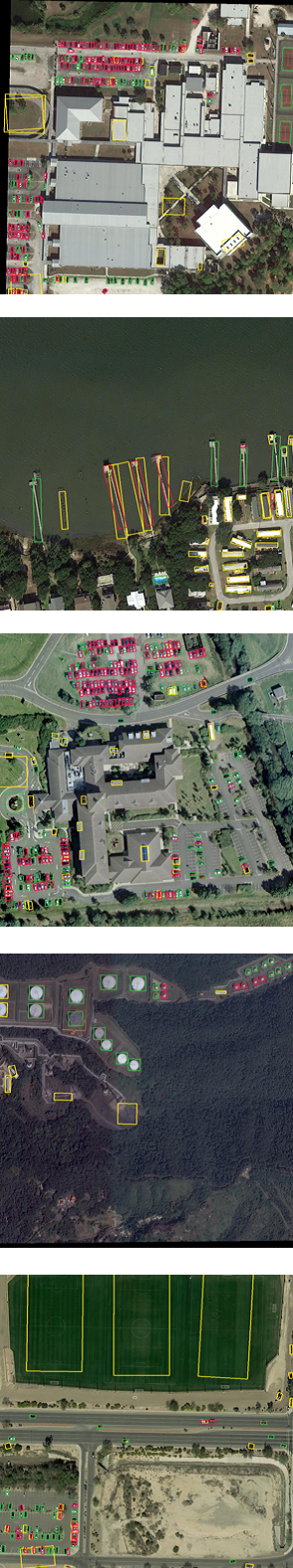}}
	\subfigure[Pseco~\cite{li2022pseco}]{
		\includegraphics[width=0.175\textwidth]{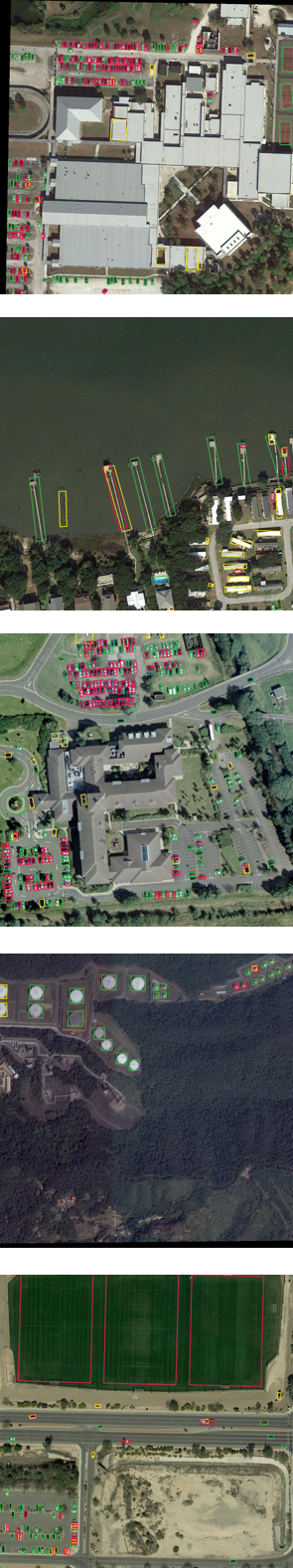}}
	\subfigure[S$^3$OD(Ours)]{
		\includegraphics[width=0.175\textwidth]{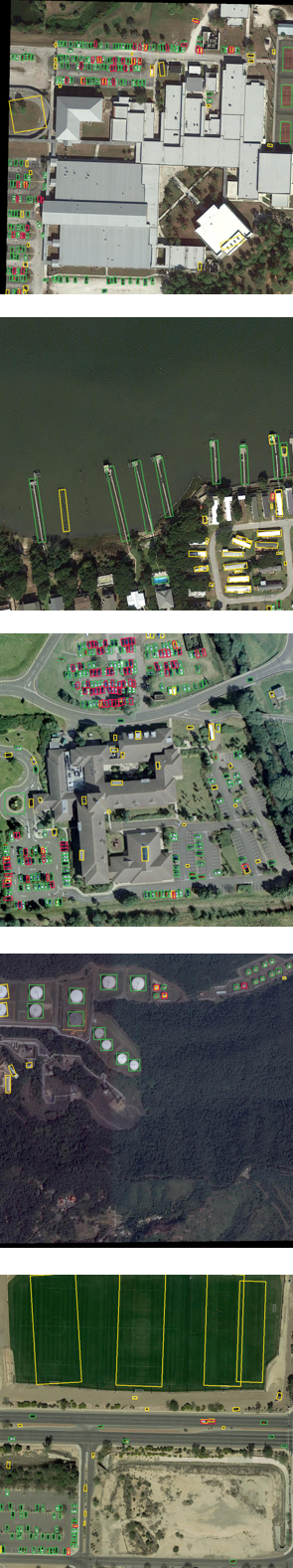}}
	\caption{Visualization of the detection results of different SSOD methods. Correct predictions are marked with green boxes, false positive predictions are marked with yellow boxes, and missing targets are marked with red boxes.
	}
	\label{fig:res}
\end{figure*}

\subsection{Main Results}
\subsubsection{Quantitative Analysis}
We compare our proposed S$^3$OD with existing SOTA SSOD methods on the DOTA-v1.5 dataset, including STAC~\cite{sohn2020simple}, Unbiased Teacher~\cite{liu2021unbiased}, Soft-teacher~\cite{xu2021end}, and PseCo~\cite{li2022pseco}. Additionally, we use the base framework without the three modules proposed in our method as an SSOD baseline. We re-implement the SOTA SSOD method for rotated object detection and adjust the appropriate hyperparameters to better adapt SSOD for aerial images. In the experiments, to ensure a fair comparison, we apply the same weak/strong augmentation strategy for all experiments. The results are shown in Tab.~\ref{tab:res}.

For partially labeled data, our proposed S$^3$OD method achieves the best performance under the training of 1$\%$, 5$\%$, and 10$\%$ labeling rate, reaching 40.02 mAP, 54.88 mAP, and 60.24 mAP respectively. This outperforms the supervised baseline by 10.72 points, 7.8 points, and 5.88 points, respectively. Similarly, our method also surpasses the SOTA method PseCo~\cite{li2022pseco} by 2.42 points, 1.90 points, and 1.98 points at different labeling rates. For fully labeled data, S$^3$OD also achieves optimal performance, exceeding the supervised baseline by 3.7 points to 67.3 mAP.
Considering the above results, our method outperforms the SOTAs by a large margin at all labeling rates on aerial images. The outstanding performance also demonstrates the superiority of our method in semi-supervised object detection tasks for aerial images.

To further demonstrate our method's effectiveness on small objects, we select several representative categories from aerial images, including ship (SP), plane (PL), small-vehicle (SV), large-vehicle (LV), bridge (BR), storage-tank (ST), and harbor (HB). In order to thoroughly validate the improvement of semi-supervised learning with limited labeled data, we present the results of S$^3$OD and other SOTA methods for each category at a labeling rate of 1$\%$, as shown in Tab.~\ref{tab:cls} (Reproducibility experiments generally exhibit similar trends, so we randomly select one fold for demonstration). 
From the results, our method shows significant improvements compared to supervised detection results in all categories. Among the categories with the highest proportions in the entire dataset, including ``ship'', ``small-vehicle'', and ``large-vehicle'', our method achieves improvements of 16.4$\%$, 75.7$\%$, and 30.7$\%$ respectively. Focusing on small objects, ``ships'', ``small-vehicle'', and ``storage-tank'' are mostly in small object sizes. Our method outperforms other SSOD methods in these three categories. Additionally, ``small-vehicle'' is the category with the most instances and the highest number of small objects in aerial images. Our method shows significant improvement in this category. In contrast, other SSOD methods show minimal improvement in ``small-vehicle'' detection, and the performance of the baseline SSOD is even lower than that of supervised learning. This confirms our previous observation that generic semi-supervised detection methods are inclined to focus on large objects while neglecting the learning of small objects, thus affecting the ability to detect small objects.

\subsubsection{Qualitative Analysis}
Fig.~\ref{fig:res} presents the qualitative comparison between S$^3$OD and other SOTA methods. We can see that, by introducing the proposed strategies, false negatives (red boxes) of small objects are remarkably reduced, indicating that small objects are more sufficiently learned. Furthermore, by focusing on the number of yellow boxes in the image, it can be noticed that compared to other SOTA SSOD methods, our approach demonstrates a stronger suppression effect on false alarms, which is the desired effect of the proposed TNL (Teacher-guided Negative Labeling) technique.

\subsection{Ablation Study}

In this section, we conduct detailed ablations to validate our key designs. Without losing generality, all the ablation experiments are performed on the single data fold with 1$\%$ labeling rate.

\begin{table}[t!]%
    \centering
	\caption{Component analysis of the proposed method}
	\label{tab:component}
	\resizebox{0.38\textwidth}{!}{%
		\begin{tabular}{@{}ccccc@{}}
			\toprule
			\multicolumn{1}{l}{} & SAT & SLA &TNL  &{mAP} \\ \midrule
			SSOD baseline      &          &          &                 & 36.0                  \\
    		\midrule
		\multirow{6}{*}{\makecell[c]{Different \\ Configurations}}  
			& \checkmark     &     &           & 39.6      \\ 
			&          & \checkmark          &                 &  37.8                  \\
			&        &       & \checkmark               & 38.1                  \\
			& \checkmark       &\checkmark       &                & 40.1                  \\
			& \checkmark         &      & \checkmark               & 40.4                \\
			&        &\checkmark       & \checkmark               & 40.4                  \\
    		\midrule
			\textbf{S$^3$OD (Ours)}  & \checkmark   & \checkmark          & \checkmark       & \textbf{41.9}                  \\
   \bottomrule
		\end{tabular}%
	}
\end{table}

\subsubsection{Component Analysis}
To verify the effectiveness of each proposed strategy individually, we conduct experiments with all possible combinations of the proposed three strategies. As depicted in Tab.~\ref{tab:component}, the baseline SSOD algorithm achieves an mAP of 36.0 when no additional strategies are employed. When we apply any of the proposed strategies, the baseline performance is consistently improved. By gradually incorporating all three strategies, the mAP shows a progress improvement, verifying each design’s effectiveness. Notably, SAT contributes significantly to the improvements. SAT enhances the baseline performance by 3.6 points because it directly impacts the supervision of small objects. These findings indirectly affirm the substantial impact of small objects on aerial image SSOD performance. Fortunately, our proposed methods effectively mitigate this impact, making them highly suitable for SSOD tasks in aerial images.

\begin{table}[t!]\tiny
    \centering
	\caption{Performance of SSOD with different thresholds used for selecting pseudo-labels}
	\label{tab:threshold}
	\resizebox{0.40\textwidth}{!}{%
    	\begin{tabular}{@{}c|c|c@{}}
    		\toprule
            Methods               & Setting            & mAP \\
    		\midrule
    		\multirow{4}{*}{\makecell[c]{Fixed\\Threshold}}               & 0.7            & 33.6 \\
    		& 0.8            & 36.6 \\
    		& 0.9            & 36.0 \\
    		& Large-0.9 $\&$ Small-0.8            & 38.1 \\
    		\midrule
    		\multirow{2}{*}{\makecell[c]{Adaptive\\Threshold}}& Class-aware          & 35.2          \\
    		& Size-aware (Ours)          & \textbf{39.6}          \\
    	    \bottomrule
    	\end{tabular}%
	}
\end{table}

\begin{table}[t!]\tiny
    \centering
	\caption{Effects of different percent numbers used in SAT}
	\label{tab:p}
	\resizebox{0.46\textwidth}{!}{%
    	\begin{tabular}{@{}c|cccccc@{}}
    		\toprule
            $P$($\%$)           & 25     & 30 &35 &40 &45 &50       \\
    		\midrule
    	  mAP         & 38.7    & 39.4 &\textbf{39.6} & 38.4 &37.8 &34.0 \\
    	    \bottomrule
    	\end{tabular}%
	}
\end{table}

\begin{figure}[!t]
\centering
\includegraphics[width=0.46\textwidth]{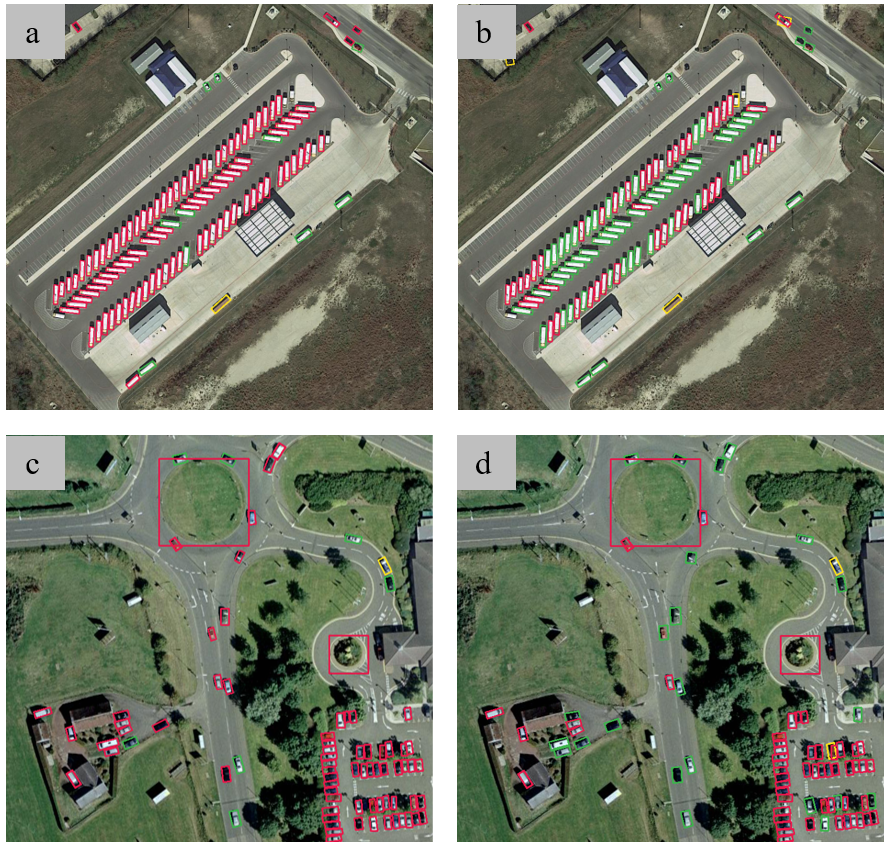}
\caption{Visualization of pseudo-labels selection. (a) and (c) are under the fixed threshold of 0.9, (b) and (d) are under our SAT.}
\label{fig.6}
\end{figure}

\subsubsection{Comparisons of Different Thresholds}

For the threshold of selecting pseudo-labels, we compare several ways of setting the threshold. These include a fixed threshold of 0.9 (baseline), a fixed threshold of 0.8 and 0.7, setting thresholds based on the object scale where a threshold of 0.9 is used for large objects and 0.8 for small objects, setting thresholds based on category distribution, and the proposed SAT method in this paper. Results are listed in Tab.~\ref{tab:threshold}. Compared to a completely fixed threshold, setting different thresholds for large and small objects shows higher performance, indicating that the optimal threshold for pseudo-labels is different for small and large objects. The proposed SAT method still improves performance, demonstrating that adapting the threshold based on the distribution is a superior strategy. Additionally, compared to setting thresholds separately for large and small objects, the proposed SAT method only requires adjusting one hyperparameter, the percentile value $P$ used in SAT. Tab.~\ref{tab:p} also presents ablation experiments with different percentile values, showing that the optimal performance is achieved when using the 35th percentile of the distribution.

Furthermore, we provide visualizations of the selected pseudo-labels when training under different strategies in Fig.~\ref{fig.6}. In this figure, green boxes denote the correct pseudo-labels, yellow boxes denote the false pseudo-labels, and red boxes denote objects missing to be pseudo-labeled. It can be observed that when using a fixed threshold, a large number of small objects are filtered out, while our method can retain more accurate pseudo-labels for small objects.

\vspace{0.3em}
\subsubsection{Different compositions in SLA}
Here, we dissect different compositions in SLA. Results are shown in Tab.~\ref{tab:sla}. When the distribution-based re-sampling strategy is used to replace the original IoU+Threhold strategy, we can achieve an improvement of 0.5 mAP. However, this improvement is relatively modest due to the limited supervision information for small objects in pseudo-labels. On the other side, when we incorporate the re-weighting strategy, which includes training loss for both large and small objects (SLA), we can achieve a more notable improvement of 1.8 mAP. This verifies that re-weighting can partially mitigate the issue of imbalanced training samples for different-sized objects during the training process. Additionally, we conduct tests on the optimal value of the hyper-parameter $K$ within the complete S$^3$OD framework, as presented in Tab.~\ref{tab:k}. The best performance is achieved when $K=2$.

\begin{table}[t!]
    \tiny
    \centering
	\caption{Component analysis of the SLA}
	\label{tab:sla}
	\resizebox{0.26\textwidth}{!}{
		\begin{tabular}{@{}ccc@{}}
			\toprule
			\multicolumn{1}{l}{} & Setting  &{mAP} \\ \midrule

		\uppercase\expandafter{\romannumeral1} & Baseline       & 36.0      \\ 
		\uppercase\expandafter{\romannumeral2}	& Resampling    &  36.5    \\
		\uppercase\expandafter{\romannumeral4}  & SLA     & \textbf{37.8}                  \\
   \bottomrule
		\end{tabular}
	}
\end{table}

\begin{table}[t!]\tiny
    \centering
	\caption{Effects of parameters $K$}
	\label{tab:k}
	\resizebox{0.38\textwidth}{!}{
    	\begin{tabular}{@{}c|cccc@{}}
    		\toprule
            $K$               & 1 &2 &3 &4        \\
    		\midrule
    	  mAP            & 39.8 &\textbf{41.9} & 41.1 &40.7 \\
    	    \bottomrule
    	\end{tabular}
	}
\end{table}

\begin{table}[t!]\tiny
    \centering
	\caption{Component analysis of the negative learning}
	\label{tab:tnl}
	\resizebox{0.40\textwidth}{!}{
		\begin{tabular}{@{}cccc@{}}
			\toprule
			\multicolumn{1}{l}{} & Setting  &FA  &{mAP} \\ \midrule

		\uppercase\expandafter{\romannumeral1} & T-BG     & 0.832     & 37.7      \\ 
		\uppercase\expandafter{\romannumeral2}	& T-BG$\_$reweight     &  0.759            &  39.5        \\
		\uppercase\expandafter{\romannumeral3}	& T-BG$\_$HNS   &   0.747            & 41.2                 \\
		\uppercase\expandafter{\romannumeral4}  & TNL (Ours)   &  0.652        & \textbf{41.9}                  \\
   \bottomrule
		\end{tabular}
	}
\end{table}

\vspace{0.3em}
\subsubsection{Different strategies of Negative learning}

For the selection of negative samples, we also compare several different strategies, including:
\begin{itemize}
    \item T-BG: Only using the teacher model to select negative samples with high background confidence scores (higher than 0.7).
    \item T-BG$\_$reweight: Following the strategy in Soft-teacher~\cite{xu2021end}, use the background confidence scores to reweight the loss of negative samples.
    \item T-BG$\_$HNS: Under the setting of T-BG, incorporate low-confidence predictions of the teacher model directly as hard negative samples.
    \item TNL: Under the setting of T-BG, incorporate the weighted low-confidence predictions of the teacher model as hard negative samples.
\end{itemize}
Here, we also calculate the overall false alarm~(FA) of the detection results of the validation set under each setting, which reflects the model's ability to discriminate negative samples. A high false alarm represents that the detector struggles to differentiate negative samples, indicating insufficient classification performance. We get $FA = FP/P$, where $FP$ represents the total number of incorrect detections, $P$ represents the number of detected objects and objects with $IoU < 0.5$ between the prediction and ground truth were classified as false positives. From the results of Tab.~\ref{tab:tnl}, we can observe that directly using T-BG to select negative samples leads to a boom in the false alarm rate. The increase in false alarms accumulates more errors in the network, resulting in performance degradation. Utilizing T-BG to reweight the negative samples can partially retain the effect of false negative samples and improve performance. Our method effectively utilizes ambiguous predictions generated by the teacher model as challenging negative samples, leading to better performance improvement. Compared to directly introducing these hard negative samples, using confidence score weighting can obtain better results. This is because, among these low-confidence predictions, a small number of positive samples may also exist. Weighting them can reduce the impact of these potential positive samples and avoid confusion between positive and negative samples by the network.

\section{Discussion}
\begin{figure}[!t]
\centering
\includegraphics[width=3.5in]{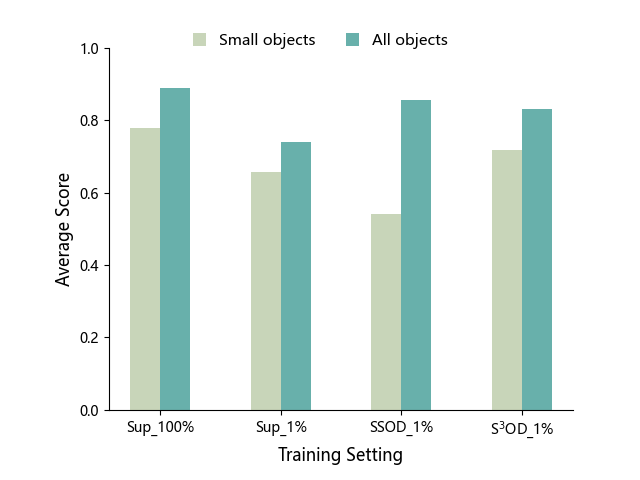}
\caption{Comparing the average confidence scores of all objects and small objects in correct predictions under the three settings of supervised by 100$\%$ labeled data, supervised, semi-supervised, and our S$^3$OD by 1$\%$ labeled data.}
\label{fig.7}
\end{figure}
Overall, extensive experiments have convincingly demonstrated the effectiveness of our proposed S$^3$OD. The detection mAP for specific categories and the corresponding visualizations also provide sufficient evidence that the S$^3$OD can optimize the detection performance of small objects on semi-supervised object detection in aerial images. To further support this point, we present the small objects' average predicted scores of different methods after semi-supervised training, which is in Fig.~\ref{fig.7}. It is clear that, in comparison to existing SSOD methods that overlook small object information in unlabeled images and solely enhance the detection of large objects, our S$^3$OD method achieves a more balanced improvement in the detection performance across all scales of objects. This, in turn, validates that our method mitigates the impact of small objects on semi-supervised object detection in aerial images.

However, apart from numerous small objects, we observe that some objects with extreme characteristics also have poor performance, such as objects with elongated aspect ratios, including `bridge' and `harbor'. Based on the detection results for the `Harbor'~(HB) category in Tab.~\ref{tab:cls}, it is evident that the SSOD method does not exhibit substantial enhancements for harbors, and our method performs similarly to existing methods when dealing with these objects. The underlying reason is that objects with these characteristics face a disadvantaged position even in detection, compared to objects with regular shapes. Similarly, these objects might be disregarded during semi-supervised training. And it is equally important to adopt differentiated criteria and select more suitable pseudo-label supervision information for disadvantaged objects with other extreme characteristics. Furthermore, aerial image object detection also faces an inherent imbalance issue, known as class imbalance. Compared to natural scenes, aerial image objects often exhibit a more severe long-tail distribution, and the class imbalance is further amplified in the process of semi-supervised detection. Categories with a larger number of samples tend to show better detection performance, allowing for the acquisition of more reliable pseudo-labels for supervision. In contrast, categories with fewer samples are more prone to being overlooked within this virtuous cycle. How to effectively mitigate these biases within the semi-supervised pipeline, will be further explored in our future work.

\section{Conclusion}
In this paper, we analyze the key factors influencing the performance of SSOD for aerial images and identify that existing SSOD methods tend to overlook the presence of small objects, which constitute a significant portion of aerial images. To address this challenge, we design a novel framework S$^3$OD to tackle the various imbalance challenges caused by small objects. In S$^3$OD, we adopt an adaptive approach to select scale-sensitive thresholds, allowing us to retain more supervision information for small objects during the pseudo-label generation stage. To mitigate the imbalance problem resulting from limited supervision, we introduce a label assignment strategy with Gaussian-based sampling and size-aware re-weighting, ensuring a more equitable assignment of positive samples for objects of different scales. Furthermore, we leverage the ambiguous predictions generated by the teacher model to enhance the model's ability to learn from challenging negative samples. Through integrating the three designs, our S$^3$OD framework presents a significant leap forward compared to the current SOTA SSOD methods for aerial images.

\bibliographystyle{IEEEtran}
\small
\bibliography{IEEEabrv,IEEEexample}

\end{document}